\documentclass{article}

\usepackage{microtype}
\usepackage{graphicx}
\usepackage{subfigure}
\usepackage{booktabs} % for professional tables
\usepackage{epstopdf}
\usepackage{url}
\usepackage{textcomp}
\usepackage{algorithmic}
\usepackage{algorithm}
\usepackage{bm}
\usepackage{amssymb}
\usepackage{pdfrender}

\usepackage{hyperref}

\usepackage[accepted]{icml2019}

\icmltitlerunning{Dream Distillation}

\begin{document}

\twocolumn[
\icmltitle{Dream Distillation: A Data-Independent Model Compression Framework}



\icmlsetsymbol{equal}{*}

\begin{icmlauthorlist}
\icmlauthor{Kartikeya Bhardwaj}{to}
\icmlauthor{Naveen Suda}{goo}
\icmlauthor{Radu Marculescu}{to}
\end{icmlauthorlist}
%
%\icmlaffiliation{*}{Work done while the author was an intern at Arm Inc., San Jose, CA, USA}
\icmlaffiliation{to}{Department of Electrical and Computer Engineering, Carnegie Mellon University, Pittsburgh, PA, USA}
\icmlaffiliation{goo}{Arm Inc., San Jose, CA, USA}
\icmlcorrespondingauthor{Kartikeya Bhardwaj}{kbhardwa@andrew.cmu.edu}
%\icmlcorrespondingauthor{\ }{\ }

\icmlkeywords{Dream Distillation, Model Compression, Knowledge Distillation, Deep Learning}

\vskip 0.3in
]



\printAffiliationsAndNotice{} % otherwise use the standard text.

\begin{abstract}\vspace{-2mm}
Model compression is eminently suited for deploying deep learning on IoT-devices. However, existing model compression techniques rely on access to the original or some alternate dataset. In this paper, we address the model compression problem when no real data is available, \textit{e.g.}, when data is private. To this end, we propose \textit{Dream Distillation}, a data-independent model compression framework. Our experiments show that Dream Distillation can achieve $88.5\%$ accuracy on the CIFAR-10 test set without actually training on the original data! 
\end{abstract}

\section{Introduction}\vspace{-2mm}
Complex deep neural networks with millions of parameters have achieved breakthrough results for many vision, and speech recognition applications. In the IoT-era, however, the edge devices are heavily hardware-constrained. Therefore, model compression of deep networks has now emerged as an important problem. Towards this end, many state-of-the-art model compression techniques such as pruning~\cite{prune3}, quantization~\cite{quant1} and Knowledge Distillation (KD)~\cite{hintonKD, atkd} have been proposed. Pruning aims at removing redundant or useless weights from deep networks, while quantization reduces the number of bits used to represent weights and activations. On the other hand, KD trains a significantly smaller student model to mimic the outputs of a large pretrained teacher model.

The existing model compression techniques above rely on access to the original training data or some unlabeled dataset. Hence, most model compression research implicitly assumes access to the original dataset. However, for many applications, the original data may not be available due to privacy or regulatory reasons (\textit{e.g.}, private medical images, speech data, \textit{etc}.). %Indeed, training a model on a big private dataset can result in complex deep learning models. 
Consequently, the industries deploying large deep learning models at the edge must compress them \textit{without} access to any original, private, or alternate datasets\footnote{Collecting alternate datasets for model compression may not always be possible, or can be very expensive/time-consuming and, hence, infeasible.}~\cite{kd123}. Therefore, in this paper, we address the following \textbf{key question}: \textit{How can we perform model compression when the original or unlabeled data for an application is not available?} We call this problem as \textit{data-independent model compression}. 

To answer this question, we propose a new framework called \textit{Dream Distillation}. Our framework uses ideas from the field of deep network interpretability~\cite{bb} to distill the relevant knowledge from the teacher to the student, in absence of access to the original training data. Specifically, our approach consists of two steps: (\textit{i}) We first exploit a small amount of metadata and the pretrained teacher model to generate a dataset of synthetic images, and (\textit{ii}) We then use these synthetic images for KD. To this end, our \textbf{key goal} is to generate synthetic data while preserving the features from the original dataset such that the teacher can transfer the knowledge about these features to the student. This effective transfer of knowledge via synthetic data can make the student model learn characteristics about original classification problem without actually training on any real data! By allowing users to deploy a model on IoT-devices without access to the private third-party datasets, data-independent model compression techniques can truly accelerate the adoption of AI on edge devices.

%\section{Learning from Small Data}
%We now discuss the challenges arising from lack of real data at the edge. 
%For instance, data can be so scarce that we need models capable of learning more effectively from Small Data. This is where techniques such as FeatureNet~\cite{pakdd} can help us learn important insights (or extract useful features) from high-dimensional, small sample-size problems. However, a
%When Big Data is indeed available but is private, the industries trying to deploy such models on the edge cannot use the original datasets for model compression. We discuss this important case below.% with the help of our recent work on Dream Distillation.
%%We now discuss how important insights can be learned from Small Data, and then describe our recent work on how to compress a complex deep learning model when we do not have access to the original dataset. 

\vspace{-1mm}
\section{Background and Related Work}\vspace{-2mm}
We first describe KD and feature visualization, which are necessary for Dream Distillation. Then, we discuss the related work on data-independent model compression and show how our approach differs from existing works.

\vspace{-2.5mm}
\subsection{Knowledge Distillation}
\begin{figure*}[!t]
\centering
\includegraphics[width=5.1in]{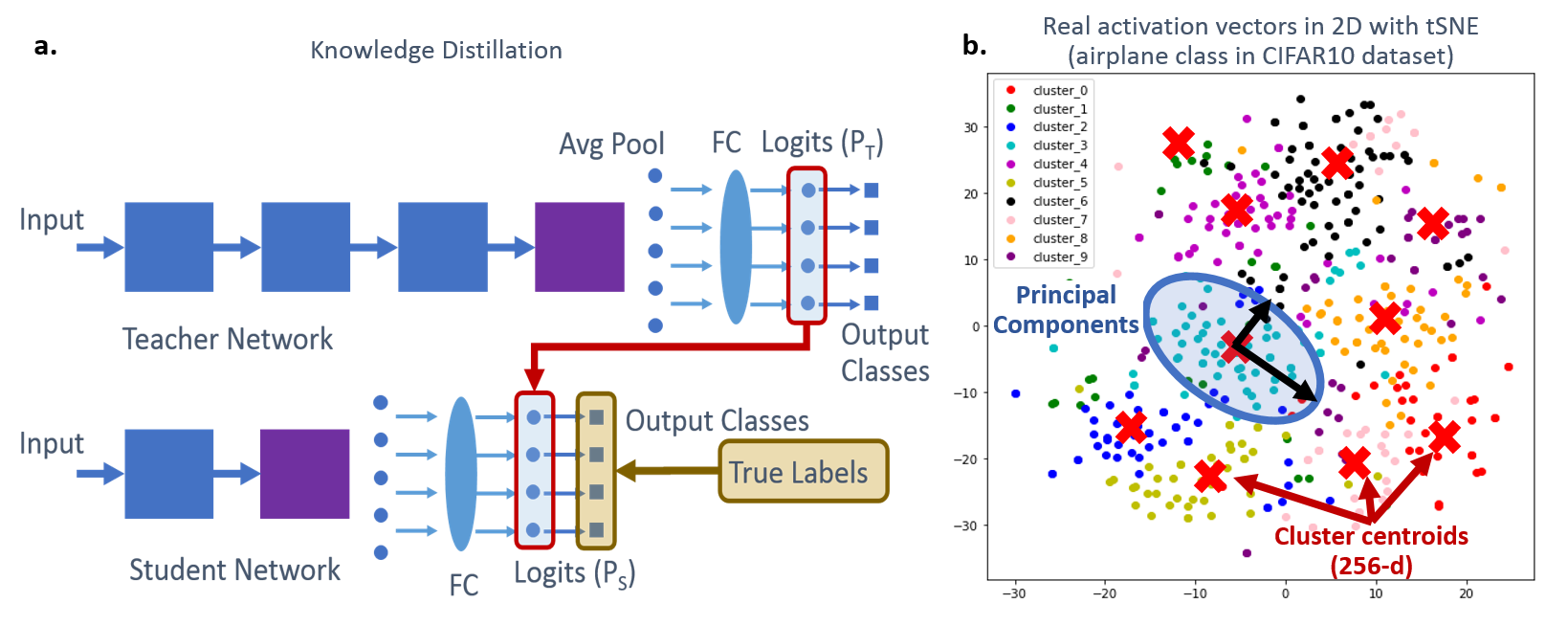}\vspace{-3mm}
	\caption{(a) Knowledge Distillation:  A significantly smaller student model mimics the outputs of a large teacher network. (b)~Metadata used in Dream Distillation.\vspace{-4mm}}
\label{kd}
\end{figure*}

\vspace{-2mm}
KD refers to the teacher-student paradigm, where the teacher model is a large deep network we want to compress~\cite{hintonKD, atkd}. In KD, we train a significantly smaller student neural network to mimic this large teacher model (see Fig.~\ref{kd}(a)). KD has also been shown to work with unlabeled datasets~\cite{mismatch}. Of note, since the term ``model compression'' usually refers to pruning and quantization, we assume KD to be a part of model compression, as it also leads to significantly compressed models.

\vspace{-2mm}
\subsection{Feature Visualization}\vspace{-2mm}
Feature visualization domain aims to visualize and understand which patterns activate various neurons in deep networks~\cite{bb}. Towards this, tools such as DeepDream~\cite{deepdream} and Tensorflow Lucid~\cite{tl} can generate an image that maximizes a given objective. For example, Tensorflow Lucid can be used to generate an image that maximally activates a given hidden unit (say, a neuron or a channel). These generated synthetic images are called as the \textit{Dreams} of a neural network. Since our work is inspired by KD and feature visualization, we call our approach \textit{Dream Distillation}.

Of note, in order to use feature visualization for data-independent model compression, we assume access to some small amount of metadata which is used to generate the synthetic images for KD. 

\vspace{-2mm}
\subsection{Data-Independent Model Compression}\vspace{-2mm}
Despite its significance, the literature on model compression in absence of real data is very sparse. A relevant prior work is \cite{kd123} where the authors propose a Data-Free KD (DFKD) framework. However, there are major differences between DFKD and the present work:\vspace{-3mm}
\begin{enumerate}
    \item DFKD requires significantly more metadata than our approach. Specifically, \cite{kd123} argue that using metadata from only the final layer under-constrains the image generation problem, and results in very poor student accuracy. Consequently, DFKD assumes access to metadata at \textit{all} layers. In contrast, Dream Distillation assumes that metadata is available only at one layer of the teacher network. Hence, in this paper, we precisely demonstrate that metadata from a \textit{single} layer is sufficient to achieve high student accuracy, something that DFKD failed to accomplish.\vspace{-2mm}
    \item When using metadata from only one layer, DFKD achieves only $68$-$77\%$ accuracy on MNIST dataset~\cite{kd123}; this means the accuracy of DFKD will be even lower for significantly more complex, \textit{natural} image classification datasets like CIFAR-10. On the other hand, we demonstrate $81$-$88\%$ accuracy on CIFAR-10 dataset without training on any real data.\vspace{-2mm}
    \item DFKD also proposes a spectral method-based metadata for synthetic image generation. However, both spectral methods and all-layer metadata can be computationally very expensive and do not scale for larger networks. Compared to these, we follow a clustering-based approach which helps generate diverse images while using significantly less computation.\vspace{-3mm}
\end{enumerate}

Finally, \cite{dfp12} focus on data-free finetuning for pruning, and show the effectiveness of their approach for fully-connected layers. In comparison, our work is much more general, as we do not focus on just the finetuning of a compressed model, but rather on training a compressed student model from scratch.

\vspace{-1mm}
\section{Proposed Dream Distillation}\vspace{-2mm}
We propose Dream Distillation to address the following research question: \textit{In absence of the original training dataset (or any alternate unlabeled datasets), how can we perform model compression without compromising on accuracy?} Specifically, for KD with teacher model trained on CIFAR-10, datasets such as CIFAR-100 and tinyImagenet have been shown to be effective alternatives for distilling relevant knowledge~\cite{mismatch}. However, since alternate datasets may not always be available, our focus here is to generate a synthetic dataset which can be just as effective at distilling knowledge as these alternate data. Hence, we assume that the alternate/original data is \textit{not} available; rather, a small amount of metadata is given for model compression.

%Consider a KD-based model compression problem where a teacher model is trained on CIFAR-10 dataset (see Fig.~\ref{flow}(a)). It has been shown that \textit{alternate datasets} such as CIFAR-100 or tinyImagenet can be used to train a student model via KD~\cite{mismatch}. The basic idea here is that during distillation, the teacher can transfer relevant knowledge about CIFAR-10 to the student even when trained with alternate datasets like CIFAR-100. The resulting student can demonstrate reasonably good accuracy on the original intended task, \textit{i.e.}, the CIFAR-10 classification. 
%%Indeed, the accuracy of student models trained on CIFAR-100 is not as high as that with real CIFAR-10 data, but is still reasonably close. 
%Since in many cases, it may not be possible to obtain even the alternate real datasets (see Footnote~2), an important question is \textit{how can we compress a deep network without using the original training set or any alternate real data, while achieving comparable accuracy?} To answer this question, we design a new framework called \textit{Dream Distillation}, a KD-based model compression technique which does not rely on access to \textit{any} real data.

\vspace{-2mm}
\subsection{Metadata}\vspace{-2mm}
We will use CIFAR-10 dataset throughout this paper. To generate our metadata, we start with the \textit{activation vectors} generated by passing $10\%$ real CIFAR-10 images through the given pretrained teacher model. The activation vectors are simply the output of average-pool layer of the teacher model (\textit{i.e.}, average output of final convolution layer; see Fig.~\ref{kd}(a)). Then, we cluster these vectors via $k$-means clustering. Finally, we use the \textit{cluster-centroids} and the orthogonal principal components (PCs) of clusters as our metadata. Fig.~\ref{kd}(b) illustrates our metadata for the airplane class for CIFAR-10 dataset and 2-D visualization of activation vector clusters. By definition, centroids refer to mean activations of clusters. Hence, using centroids reduces the privacy-concerns since we do not use activations (or any other identifying information) directly from real images.

For WRN40-4 teacher model, our metadata merely amounts to 0.58MB which is about $100\times$ smaller than the size of even $10\%$ real CIFAR-10 images (around 58MB). We next use this metadata (centroids and PCs) and teacher model to generate synthetic images. Of note, image generation techniques such as Generative Adversarial Networks (GANs) cannot be used for our problem since GANs also rely on availability of real data~\cite{gan}.%In contrast, we push the model compression problem to the limit and do not use any real data.

\vspace{-2mm}
\subsection{Dream Generation and Distillation}\vspace{-2mm}
Let $c_k$ be the centroid for cluster $k$, and $p^k_j, j \in \{1,\ldots,m\}$ denote its $m$ PCs. We first create \textit{objectives} from the metadata, and then optimize them to generate images. Specifically, we add a small noise to the centroids in the direction of PCs to generate new \textit{target activations}: $t_i = c_k + \sum_j \epsilon_j p^k_j, i \in \{1,\ldots,n\}$. Here, $n$ is the number of images to be generated, and $\epsilon_j$ is Gaussian noise for $j^{th}$ PC. To compute $\epsilon_j$, explained variance of $j^{th}$ PC is used as the variance in the Gaussian noise. Adding noise proportional to the explained variance of corresponding PCs makes our target activations mimic the real activations. 

Therefore, by adding small noise to mean activations (\textit{i.e.}, cluster-centroids), the target activations emulate the behavior of real samples at teacher's average-pool layer. Then, to generate the images, we must find an image $X_i$ whose average-pool activations are as close as possible to $t_i$. Therefore, we generate the synthetic images $X_i$ as follows:\vspace{-1mm}
\begin{equation}
\displaystyle{\min_{X_i} ||g(X_i)-t_i||^2_2 \ \ \ i \in \{1,\ldots,n\}}\vspace{-1mm}
\label{eq1}
\end{equation}
where, the function $g$ refers to the average-pool output of the teacher network. 
%We use Tensorflow Lucid tool to minimize the above objective. 
We used about $m=50$ PCs per cluster and generated a total of $n=50,000$ synthetic images for CIFAR-10 classes. To generate the synthetic images, we minimize the objective in~(\ref{eq1}) by using Adam optimizer with a learning rate of $0.05$, $\beta_1 = 0.9$, and $\beta_2=0.999$. We optimize the image $X_i$ for 500 iterations. Finally, these synthetic images are used to train the student via KD. 

The main advantage of our clustering-based approach is that it enables more diversity among the generated synthetic images and, hence, achieves high accuracy. To summarize, the idea is to generate synthetic images, and then use them to distill knowledge about real data to the student.

%For Dream Distillation, we create custom objectives from the metadata. Specifically, we first generate target activations from the metadata by adding a small amount of noise to the cluster-centroids along the directions of PCs. 
%As demonstrated in Fig.~\ref{dd}(b), a two-dimensional visualization\footnote{Two-Dimensional Visualization via tSNE: \url{https://bit.ly/2FUmCzj}} of target activations (blue triangles) and real data activations (red circles) at the average pool layer of the teacher network shows that their distributions are quite similar. 
%Next, we generate $50,000$ images by using Tensorflow Lucid with the objective that activations of generated images at teacher's average pool layer must be as close as possible to these target activations. Finally, these generated (synthetic) images are used for distilling the knowledge from teacher to the student. 

\vspace{-1mm}
\section{Experimental Setup and Results}\vspace{-2mm}
For the CIFAR-10 dataset, our teacher is a large Wide Resnet (WRN)~\cite{atkd} WRN40-4 ($8.9$M parameters, $95\%$ accuracy) model, and our student model is WRN16-1 (100K parameters). Training the WRN16-1 student via Attention Transfer KD~\cite{atkd} on WRN40-4 teacher results in $91\%$ accuracy and $89\times$ fewer parameters than the teacher. %We use four different datasets containing $50,000$ images: (\textit{i}) Random noise images (blue), (\textit{ii}) Dream Distillation images (red), (\textit{iii}) CIFAR-100 images as an alternate dataset (yellow), and (\textit{iv}) Real CIFAR-10 dataset (violet).
\begin{figure*}[]
\centering
\includegraphics[width=5.0in]{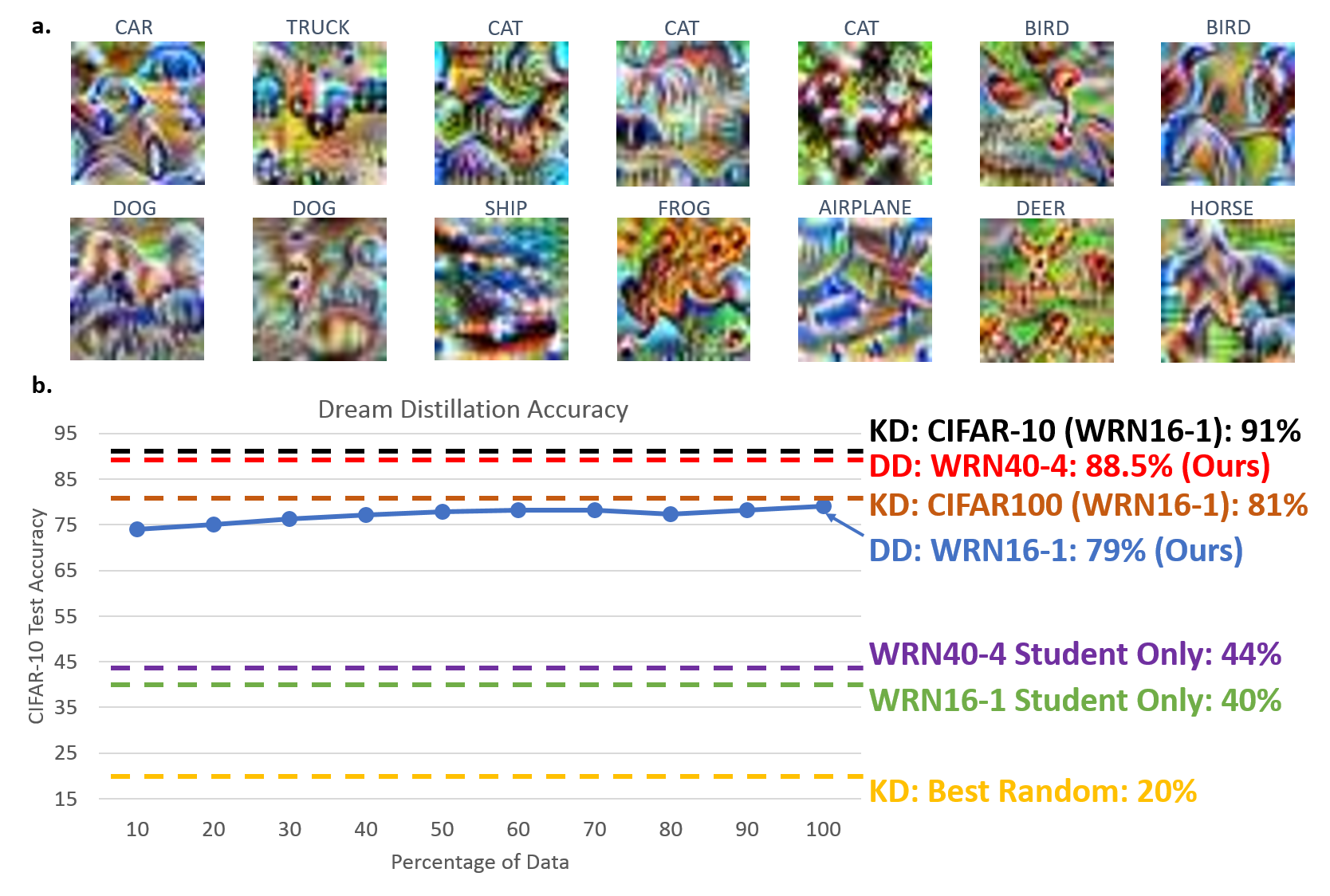}\vspace{-5mm}
	\caption{(a) Generated synthetic data for CIFAR-10 classes, and (b) Accuracy of student models trained on random, synthetic, alternate CIFAR-100, and real CIFAR-10 images.\vspace{-4mm}}
\label{ddres}
\end{figure*}

Fig.~\ref{ddres}(a) shows samples generated by our approach for different CIFAR-10 classes. As evident, for classes such as car, truck, deer, \textit{etc.}, key features like distorted-animal-faces/wheels are visible. On the other hand, images from classes such as cat, frog, ship are hard to interpret. For instance, to generate cat samples, the teacher network generally creates a striped pattern which may not be the most distinguishing feature of a cat (although many cats do have a striped pattern!). Therefore, the generated images \textit{do contain key features} learned by the teacher network for various classes (\textit{e.g.}, stripes for cats, \textit{etc.}) even though the images look far from real. Hence, these synthetic images can transfer relevant knowledge about the real data.

Next, in Fig.~\ref{ddres}(b), we compare student models trained via KD on four datasets: (\textit{i})~random noise images, (\textit{ii})~images generated via Dream Distillation, (\textit{iii})~CIFAR-100 as the alternate data, and (\textit{iv})~CIFAR-10 training set. The accuracy is reported for CIFAR-10 test set. The solid blue line shows how the accuracy of Dream Distillation varies with the number of synthetic images used for training (\textit{e.g.}, $10\%$ data means $5000$ synthetic images since we generated total $50,000$ images). Fig.~\ref{ddres}(b) demonstrates that the accuracy of Dream Distillation is comparable to that of CIFAR-100 (around $80\%$), both being around $10\%$ lower accuracy than the student trained on real CIFAR-10 data. Further, we demonstrate that the WRN40-4 student model (which is the same as the teacher) trained via Dream Distillation achieves $88.5\%$ accuracy on CIFAR-10 test set without training on \textit{any} real data! Again, for metadata available only at one layer, the prior DFKD model achieves merely $68$-$77\%$ accuracy even for MNIST dataset~\cite{kd123}. Hence, it would achieve much lower accuracy for CIFAR-10.

Finally, if these generated images are used to train a model without the teacher, WRN40-4 model achieves only $44\%$ accuracy, whereas the same model achieves $2\times$ better accuracy with the pretrained teacher model ($88.5\%$). This shows that the generated images can very effectively transfer knowledge \textit{from the teacher} to the student. Hence, the synthetic images generated via Dream Distillation can transfer significant knowledge about the real data without accessing any real or alternate datasets. This can greatly increase the scale of deep learning at the edge since industries can quickly deploy models without the need for proprietary datasets.

\vspace{-1mm}
\section{Conclusion and Future Work}\vspace{-2mm}
In this paper, we have proposed Dream Distillation, a new approach to address model compression in absence of real data. To this end, we use a small amount of metadata to generate synthetic images. Our experiments have shown that models trained via Dream Distillation can achieve up to $88.5\%$ accuracy on the CIFAR-10 test set without training on any real data. 

In future, we plan to run experiments to validate our approach on applications with medical/speech data, where different classes' representations may not be fully-separable. We will also conduct more ablation studies to analyze the impact of using more real data to generate metadata (\textit{e.g.}, instead of $10\%$ data, what if we use $20\%$ data, \textit{etc.}).

%% Could be useful for data augmentation, pruning, and quantization without real data also.

%% You could put tensorflow lucid link as footnote.

%% Emphasize somewhere that you are the first to address data-independent model compression and communication-aware model compression.

\vspace{-1mm}
\section*{Acknowledgments}\vspace{-2mm}
The authors would like to thank Dr. Liangzhen Lai (Facebook) for many useful discussions throughout this project.\vspace{-2mm}

\bibliography{example_paper}
\bibliographystyle{icml2019}


\end{document}